# Efficient support ticket resolution using Knowledge Graphs

Sherwin Varghese

sherwin.varghese@sap.com
SAP Labs India
EPIP Zone, Whitefield
Bengaluru, Karnataka 560066

James Tian

james.tian@sap.com
SAP Labs US
Newport Beach,
CA 92663

*Abstract*— **A review of over 160,000 customer cases indicates that about 90% of time is spent by the product support for solving around 10% of subset of tickets where a trivial solution may not exist. Many of these challenging cases require the support of several engineers working together within a "swarm", and some also need to go to development support as bugs. These challenging customer issues represent a major opportunity for machine learning and knowledge graph that identifies the ideal engineer / group of engineers (swarm) that can best address the solution, reducing the wait times for the customer.**

**The concrete ML task we consider here is a learning-to-rank (LTR) task that given an incident and a set of engineers currently assigned to the incident (which might be the empty set in the non-swarming context), produce a ranked list of engineers best fit to help resolve that incident. To calculate the rankings, we may consider a wide variety of input features including the incident description provided by the customer, the affected component(s), engineer ratings of their expertise, knowledge base article text written by engineers, response to customer text written by engineers, and historic swarming data.**

**The central hypothesis test is that by including a holistic set of contextual data around which cases an engineer has solved, we can significantly improve the LTR algorithm over benchmark models. The article proposes a novel approach of modeling Knowledge Graph embeddings from multiple data sources, including the swarm information. The results obtained proves that by incorporating this additional context, we can improve the recommendations significantly over traditional machine learning methods like TF-IDF.**

## I. Introduction

Customers are an integral part of the success of an organization. The ability to resolve customer cases quickly and efficiently is a priority for all businesses. However, the amount of time taken to resolve a customer ticket is often quite high due to the complexity of the issue or due to delays in identifying the engineering experts. This results in a high turnaround time for customers, affecting business outcomes. The problem of identifying the ideal engineers for a given product support issue are typically considered solving using traditional ML approaches such as TF-IDF, Random Forest and other models. These approaches are relatively easier to train and has low model complexity. However, it is not possible to capture the relationships among the engineers who teamed up to solve the problem. For most of the support incidents, the first engineer who is assigned the problem are not the final processors solving the ticket. In such scenarios, traditional ML models may lead to low accuracy.

The proposed system develops a Knowledge Graph considering the engineers that collectively solved customer incidents, the available product documentation, and the descriptions of the product components. The matching of experts to cases and swarms can be significantly improved with the inclusion of a more holistic set of inputs and far more sophisticated learning algorithms, that are trained to directly optimize the expert matching task. Historic data around who participated in which swarms and who ultimately resolved which incidents are used as both our training and evaluation data. The learning-to-rank task produces a ranked list of engineers best fit to help resolve that incident. The performance of the proposed system is evaluated against traditional ML benchmark models.

The article is divided into 5 sections. Section II describes the current state of the art and section III explains the proposed system. Benchmarking and results are provided in section IV and the section V talks about the future scope of the research.

## II. State of the Art

There is a current expert matching system which is deployed in production, although not yet integrated into the ISE. This expert matching system considers some signals such as how many incidents an engineer has solved and how many KBAs an engineer has written along with some temporal and text similarity features to produce the ranked list. This system is, as far as we are aware, not learning based. It uses a predefined set of weights, which, however, might be determined by simple linear regressions, to combine the different input features to produce its rankings. Furthermore, there are many more sophisticated features this system does not consider such as who is currently working on the incident, the graph structure of who engineers have worked with in the past, and the rich natural language present in incident communications and knowledge base articles that goes far beyond simple text similarity features.

To go beyond the currently deployed expert matching system, we look towards state of the art in two fields: NLP, specifically in natural language understanding (NLU), and graph/network theory.

In NLU, the current state of the art for generating rich representations of text is by far the set of transformer-based very large language models (VLLMs) such as BERT [1], RoBERTa [2], ALBERT [3], GPT3 [4], etc. [5-7] These VLLMs take natural language as input and produces contextual embeddings that can be used for downstream tasks such as sentence classification or named entity recognition. When end-to-end fine-tuned, these VLLMs achieve state-of-the-art performance on a wide range of tasks such as all the tasks within the GLUE [8] and Super-GLUE [9] benchmarks. The difficulty in using these models is in their size and training complexity. Generally, the biggest gains in performance are seen when these models are allowed to be end-to-end fine-tuned which means that they must be brought into the model optimization stage. This optimization can be prohibitively computationally expensive for large datasets with long form text even on specialized hardware such as GPUs and TPUs. As such, we will need to experiment with how these models can be efficiently brought into the training and inference procedure, and where it would be appropriate to use them. Significant pre-processing of the text to, e.g., highlight important snippets, might be necessary.

In graph and network theory, there are node-similarity measures (e.g., Jaccard similarity), centrality measures (e.g., page rank), and link prediction algorithms (e.g., triadic closure) that can be used to analyze graphs and help make recommendations. We may use these measures to aid or models or help form baseline models of comparison; however, we look towards more recent developments in graph neural networks (GNNs) (e.g., GraphSage [10] or PinSage [11]) and neural structured learning (NSL) [12] as feature generators for our LTR algorithms. We make this choice because GNNs and NSL create dynamic embeddings for graph nodes that depend on learned weights which are set by directly optimizing a loss function to perform best on a given task while taking the graph structure into account, via regularization of the loss function. Traditional measures of similarity, centrality, or link prediction, in contrast, are static and depend only on the graph structure. An analogous example is the relationship between a deep convolutional neural network and pre deep learning methods of feature extraction on images. Allowing a deep neural network to perform the feature extraction automatically, based only on optimization of a loss function, is generally a much more powerful approach than using manually engineered features if the data volume is large enough, which we do believe to be the case for us.

### III. PROPOSED SYSTEM

The solution comprises of defining the ontology Knowledge Graph. Some of the terminologies used in the system are explained as follows:

- **Incidents:** Incidents are the tickets or support requests that are raised by the customer or implementation partners for SAP product support.
- **Component:** Component refers to the tag that helps to identify the product / scenario where the customer is facing an issue.
- **Knowledge Base Articles (KBAs):** These are the help documentation, notes and possible resolution(s) of previous customer issues.
- **User:** This table contains the engineer information including their corporate ID and their emails.
- **Swarm:** Some complex incidents may not be solved by a single expert. For certain issues, collaboration among engineering experts is needed. This group of engineers that collectively solved an incident in the past form a swarm.

The ontology is described in Fig. 1. The key targets of interest are – engineers that were involved in an incident and engineers that belong to a swarm.

#### A. Data Sources.

The key data sources for the project are obtained from the SAP's internal ticket resolution, knowledge sharing systems. Additionally incident information and KBAs are also obtained from the Service Now data sources.

The key data sources include -
- **Infodocs:** It provides both incident-side and engineer-side data. On the incident side, it provides the original description of the incident, the steps to reproduce, the product area and the relevant component as selected by the customer. On the engineer side, infodocs provide engineer responses to customers which we can use both as input features for engineers and to inform which engineer resolved which incident. All the information contained within infodocs are not necessary. The following fields are considered for the scope of this evaluation:
    - Incident IDs – required primary key for an incident.

- - Communication summary – includes an anonymized version of all communications to/from customers and IDs of responding engineers.
  - Processor IDs (D/I user IDs) – to understand which engineer processed the incident. Will also be used in joining to other data sources.
  - Components – to understand which components are affected by the incident.
  - Incident creation date, incident confirmed date – will give temporal context to the evaluation during a specific time frame. This may also help in data composition.
- **Engineer Component Expertise:** This information will likely be helpful to benchmark component-centric methods with more holistic, learning based, methods. This data will be joined to other data sources via corporate engineer IDs. For component expertise data we require:
  - Engineer corporate ID (D/I user IDs).
  - Component lists with expertise ratings connected to those D/I user IDs.
- **Product Support Knowledge Base Articles (KBAs):** Certain incidents warrant the creation of a KBA within the database that clearly documents the problem and its resolution so that similar problems can be solved quickly and easily in the future. The authors of such KBAs can be seen as experts who are well fit to resolve such types of problems. Hence, the inclusion of KBA and authorship data provides us a rich set of contextual information on the expertise of engineers that goes well beyond their stated component expertise evaluations. For KBA data, the following fields are considered:
  - The full text of the article (including "See Also" and "Keywords"). Images and other media within the articles are out of scope as it would make the models complex.
  - Responsible user IDs.
  - Processor user IDs.
  - Category.
  - Component.
  - KBA creation date.
  - KBA ID.

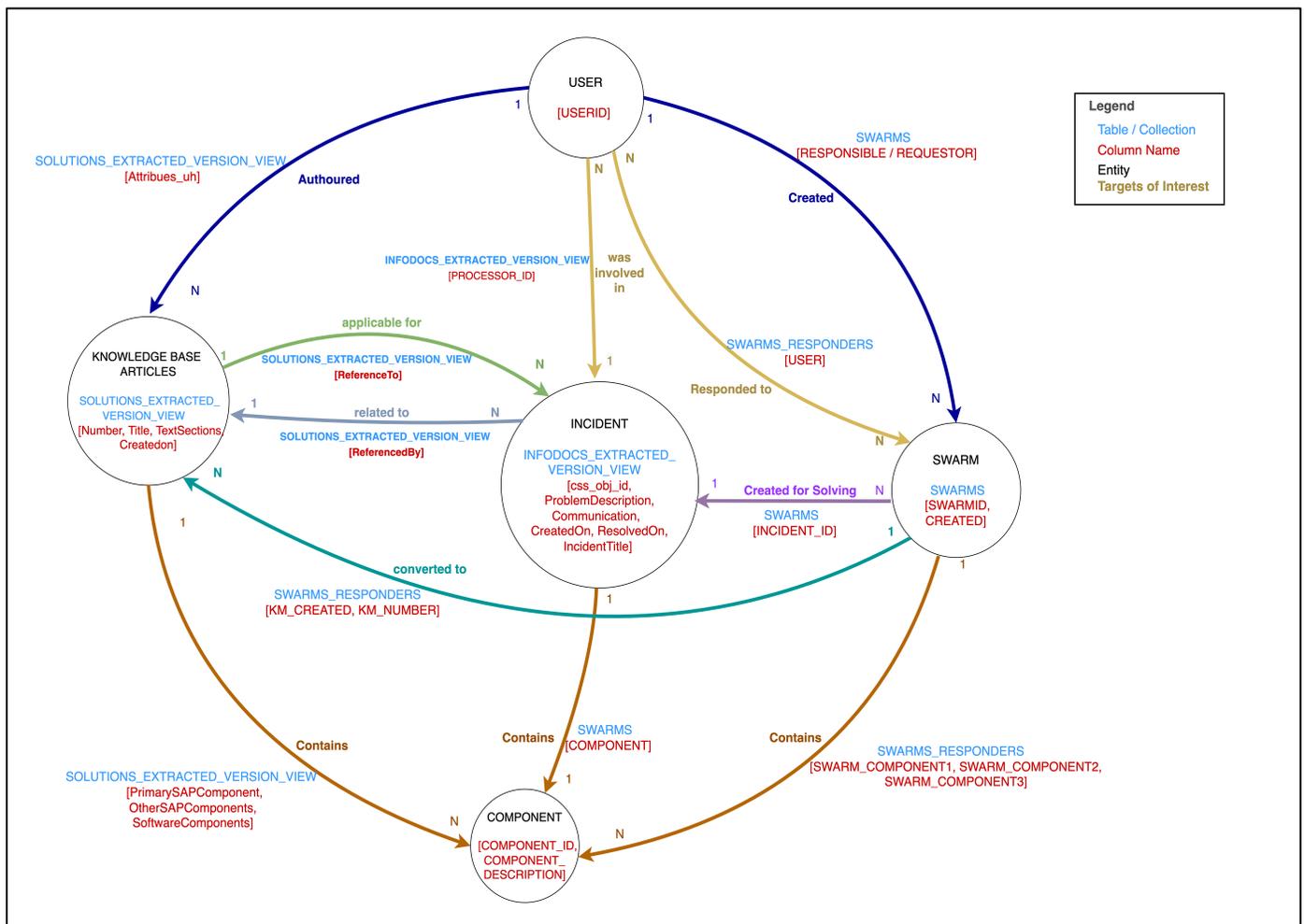

Fig. 1. Knowledge Graph Ontology of the proposed system that combines multiple relational and non-relational data sources.

- **Swarming data:** This data consists of swarms that have been requested and responded to in the past. This data gives us training and evaluation data for the swarming scenario. In particular, the responding swarmer provides us a prediction target. For this data we require:
  - Incident ID to connect the swarm to an incident.
  - User ID of the swarm requestor.
  - User ID of the swarm responder.
  - Swarm ID in case multiple swarms are initiated for a single incident.
  - KBA IDs created for the swarm.
  - Swarm creation date.
  - Component.
  - Swarm component.

With these data sources aggregated, a graph network is constructed which incorporates and joins all the data so that graph-based approaches can be leveraged.

*B. Approach and Architecture of the system.*

The problem of expert matching and swarming uses Deep Learning models to ingest two basic types of data: graph/network structured data and unstructured natural language data. Specifically, we leverage neural methods, graph neural networks (GNNs) for the former and transformers, recurrent neural networks (RNNs) for the latter, which are powerful parametric models that can be trained to directly optimize for the recommendation task.

Concretely, Engineers, KBAs, and Incidents are treated as vertices on a graph which connect, via edges, to other vertices as appropriate. For example, an engineer should be connected to incidents they have solved, KBAs they have written, and other engineers who they have swarmed with. Natural language understanding (NLU) (including neural networks) is used to ingest the natural language features of incidents and KBAs and to create input embeddings (i.e., a set of numbers that encode the relevant information) for the graph neural network which will sit on top. A simple one-hot embedding (a 1 indicates a relevant component is present and 0 is used for all other components) can be used to encode component information. The GNN then takes all relevant input embeddings and the underlying graph structure to produce a final set of embeddings for each engineer in the graph.

The initial phase involves data cleaning for the KBA, Historical Incident Communication, Component description. The first task for any data processing pipeline is of course extraction. For stage 1, the necessary data sets are extracted from their corresponding databases and joined together. Next stage involves two types of pre-processing – creating the inputs for the NLU pipeline and transforming the data into a graph-structure for the graph methods.

Fig. 2. depicts the architecture of the proposed system. Creating inputs for the NLU pipeline involves significant data cleaning – likely using regex and the like, since the Infodocs is one giant text blob with all communications concatenated with each other. Significant data processing will need to be done to separate out the communications and connect them back to the engineers as appropriate.

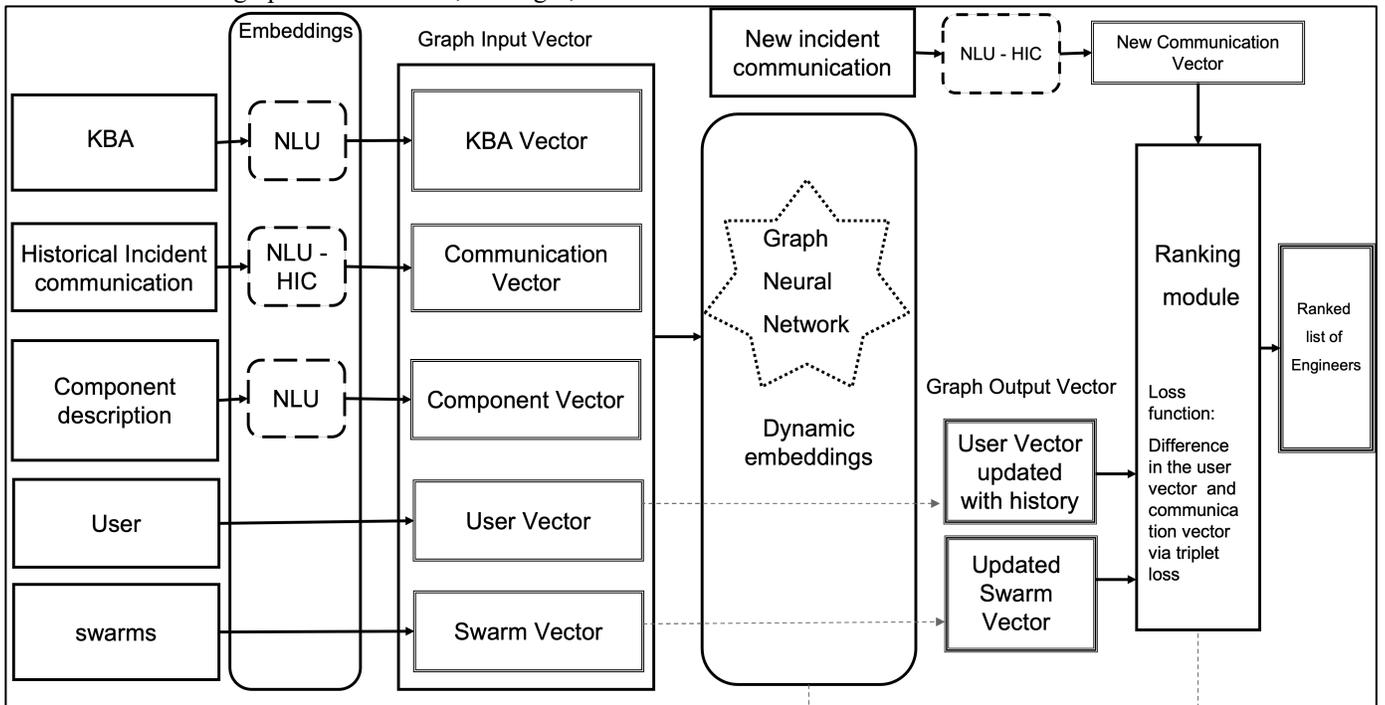

Fig. 2. The architecture of the proposed system.

5Furthermore, non-VLLM based NLU using lightweight ML models, such as simple NER, sentence classification, etc., may also have to be performed to make the data volume computationally feasible for VLLMs to process.

To transform the data into a graph-structure, we need to extract the relevant relationships inherent within the data. For example, two engineers who worked on the same incident would be linked with each other as they belong to the same swarm. We will have to determine how the edges in the graph should be defined exactly given that a naïve implementation may lead to perhaps a too densely or too sparsely connected graph. Furthermore, we noted that the KBAs are hyperlinked to each other which suggest that there's even richer graph structure inherent in the KBAs that can be exploited.

After the data processing phase, we approach this problem by building models on top of the NLU models and the Knowledge Graph model. Specifically, as mentioned in the State-of-the-Art section, we leverage neural methods, PinSage, and VLLMs for the natural language data. Our models are powerful parametric models (i.e. they rely on a set of learned parameters) that can be trained to directly optimize for the task we are considering.

The data sets that contain natural language includes KBAs, Incident Communication and the Component Description. This would need NLU processing to generate embeddings. The embeddings are then converted into their corresponding vectors. NetworkX is used to model the Knowledge Graph.

Once the KG is generated, when a new incident is encountered by the system, the Incident details are converted to a vector using NLU techniques. The incident may or may not be assigned to a swarm. If the incident is assigned to a swarm, the initial context can be obtained from engineers and by querying the graph using incident embeddings that were previously worked on will help in ranking the engineers. If the incident is not assigned to a swarm yet, the incident vector is compared to the KBA embeddings, historical incident communication to identify the swarm of engineering experts that can best solve the incident.

The Ranking Module obtains the updated user vector with historical communication and the updated swarm vector, and it uses a triplet loss function to compute the difference in the vectors of the incoming incident and generates a ranked list of engineers. An implementation based on PinSage [11] algorithm is used within the ranking module. Since, the total number of engineering experts are almost constant compared to the incoming incidents, a ranking all the engineers can be done with minimal memory resources.

Next, the incident for which we are finding recommendations for through our NLU pipeline, get the engineer embeddings for engineers already working on the incident in the case of swarming from our GNN, and use the combination of these two embeddings to query our graph for the most similar embeddings of other engineers. We can train this entire system in an end-to-end fashion using negative sampling methods.

After our AI model is trained, it is queried using validation and testing data sets, the recommendations are matched to the target data for evaluation.

## IV. EVALUATION AND RESULTS

### A. Benchmarking models and criteria.

As the system provides a ranked list of engineers given a new incident, the top-k hit ratio is an ideal metric to evaluate the performance of the models. Top k hit ratio refers to a number k and if the engineer gets listed within the $k^{th}$ number, it is considered as a hit.

Traditional models including TF-IDF, XGBoost and Random Forest were used to benchmarking the proposed deep learning-based system. For the traditional NLP approaches, the corpus was limited to the collection of databases – Infodocs, KBA, Swarms and Component descriptions. The dataset was sampled randomly into 2 sets – one set containing 10K examples and the other set containing 100K examples. For each of these sets, the top-50, top-100 and top-200 hit ratios were measured. Additionally, the timeline analysis were limited to 2019 and 2020, since there were limited "swarm" record information before this period.

During 2019, there were 781,083 records that have an assignable processor user ID. Out of these, 678,047 records contain the top 5k most prolific users. In 2020, there were 1,396,463 records that had assignable processor user ID and 1,061,330 records with the top 5k

---

**Algorithm: Generate Graph Neural Network**

1:     **procedure** generateGNN
2:         **ETL** on KBA, Communication, Component, User and Swarm
3:         perform NLU transformations for KBA, Communication, Components
4:         Normalize embeddings into vectors
5:         Model the Knowledge Graph using the vector embeddings as a NetworkX graph
6:         Ranking Module uses PinSage implementation, generates the vectors for the new incident
7:         Rank the engineers based on the triple loss function result
8:     **end procedure**

| | 10K Examples | | |
|---|---|---|---|
| **ML Model** | **Top 50** | **Top 100** | **Top 200** |
| TF-IDF | 0.48 | 0.58 | 0.68 |
| Random Forest | 0.0065 | 0.015 | 0.043 |
| XGBoost | 0.011 | 0.023 | 0.101 |
| Knowledge Graph with PinSage and embeddings | 0.64 | 0.77 | 0.85 |
| | 100K Examples | | |
| **ML Model** | **Top 50** | **Top 100** | **Top 200** |
| TF-IDF | 0.35 | 0.59 | 0.55 |
| Random Forest | 0.007 | 0.012 | 0.02 |
| XGBoost | 0.01 | 0.014 | 0.021 |
| Knowledge Graph with PinSage and embeddings | 0.70 | 0.65 | 0.78 |

Table 1. Top k hit ratio for the system.

most prolific users.

### B. Results

The summary of the results is described in Table 1.
The first benchmark model uses TF-IDF and cosine similarity of TF-IDF vectors to measure the similarity of the incidents. The results were used as the base scores for the subsequent models. Random Forest and XGBoost models were chosen as the traditional ML models.

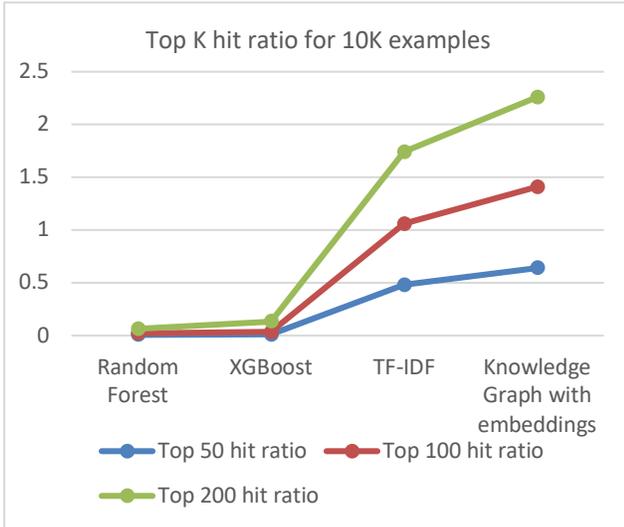

Fig. 3. Top k hit ratio for the system for 10K samples.

However, the results from these models were very much less than TF-IDF. The proposed system was then trained and benchmarked. The training time and the GPU resources of the proposed system was significantly higher than all the other models. However, the results showed a significant improvement in the top-50, top-100 and top-200 hit ratio consistently.

The following graphs (Fig. 3 and Fig. 4) depict the summary of the results obtained. Overall, the proposed system outperforms the other ML models by a huge

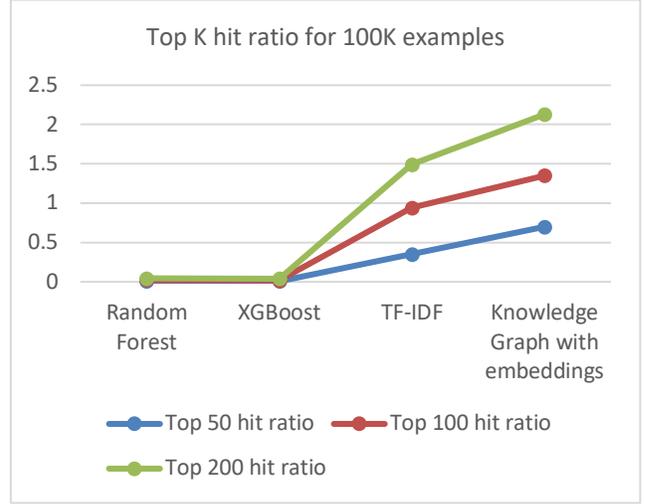

Fig. 4. Top k hit ratio for the system for 100K samples.

margin.

### V. CONCLUSION AND FUTURE SCOPE

This article proposes the design aspects of efficient ticket / incident resolution system that uses Graph Neural Networks and embeddings to efficiently rank engineers. The proposed system uses a novel approach of building a Knowledge Graph using embeddings from VLLMs and using the GNN algorithms to rank engineers based on their previous experiences of solving similar issues, incorporating their domain knowledge. The results provide a promising outlook for the proposed system when compared with traditional ML approaches. The Knowledge Graph on embeddings provides additional context by representing data relationships better.

The future scope would include incorporating employee time zones and their calendar to suggest the expert and further reduce delays in ticket resolution. Additionally, updating and deploying the model as a service based on updated KBAs, incidents and user would be a future direction. Intranet crawling may be necessary to establish links between KBAs. Hence, developing efficient crawlers can add meaningful information to the Knowledge Graph and will be considered in the future.

### VI. ACKNOWLEDGEMENTS

This work is achieved with the support and resources from the SAP New Ventures and Technologies team.